\documentclass[runningheads]{llncs}
\usepackage{CJKutf8}
\usepackage[T1]{fontenc}
\usepackage{graphicx}
\usepackage[misc]{ifsym}
\usepackage{bbding}
\usepackage{dsfont}
\usepackage{amsmath, amssymb, booktabs, float}
\usepackage[colorlinks,
linkcolor=blue,
anchorcolor=blue,
citecolor=blue]{hyperref}

\begin{document}
\begin{CJK}{UTF8}{gbsn}
%
%\title{Flexible Registration Network: Collaborative Division of Labor among Encoders Makes Cross-domain Registration Better}
\title{Encoding Matching Criteria for Cross-domain Deformable Image Registration}
\titlerunning{Cross-domain Deformable Registration}
\author{Zhuoyuan Wang \and Haiqiao Wang \and Yi Wang\Envelope}
\authorrunning{Z. Wang et al.}
\institute{
	Smart Medical Imaging, Learning and Engineering (SMILE) Lab,\\
	Medical UltraSound Image Computing (MUSIC) Lab,\\
	School of Biomedical Engineering,
	Shenzhen University Medical School,\\
	Shenzhen University, Shenzhen, China\\
	\email{onewang@szu.edu.cn} \\
}
\maketitle              % typeset the header of the contribution
\begin{abstract}
Most existing deep learning-based registration methods are trained on single-type images to address same-domain tasks.
However, cross-domain deformable registration remains challenging.
We argue that the tailor-made matching criteria in traditional registration methods is one of the main reason they are applicable in different domains.
Motivated by this, we devise a registration-oriented encoder to model the matching criteria of image features and structural features, which is beneficial to boost registration accuracy and adaptability.
Specifically, a general feature encoder (Encoder-G) is proposed to capture comprehensive medical image features, while a structural feature encoder (Encoder-S) is designed to encode the structural self-similarity into the global representation.
Extensive experiments on images from three different domains prove the efficacy of the proposed method.
Moreover, by updating Encoder-S using one-shot learning, our method can effectively adapt to different domains.
\textit{The code is publicly available at}
\url{https://github.com/JuliusWang-7/EncoderReg}.
 
\end{abstract}
\section{Introduction}
\begin{figure}[t]
	\centering
	\includegraphics[width=0.99\textwidth]{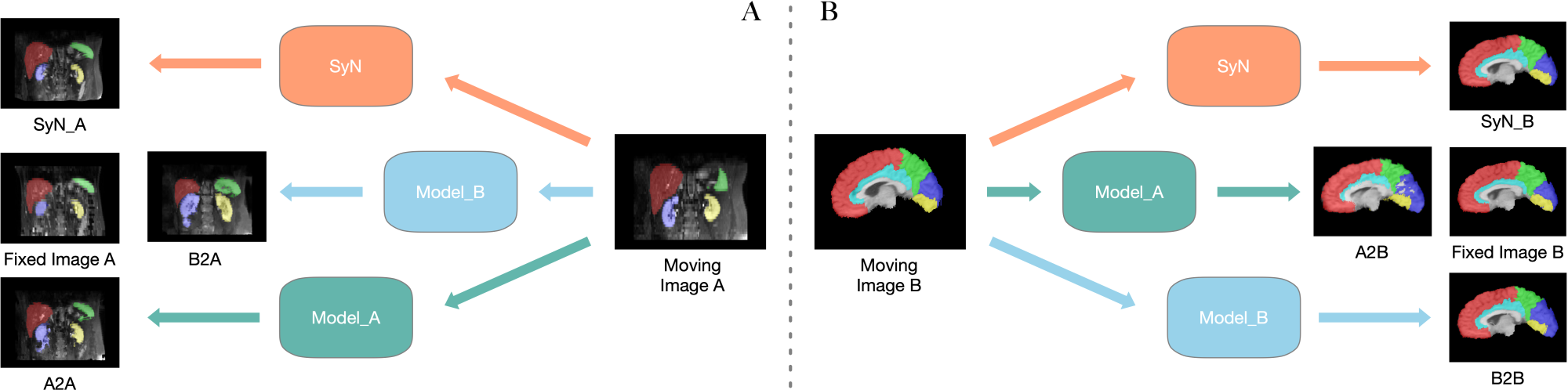}
	\caption{The registration for domain A (abdomen MRI) and domain B (brain MRI).
		Model\_A and Model\_B represent the deep learning-based registration networks trained on domain A and domain B, respectively.
		It can be observed that Model\_A and Model\_B cannot provide accurate results for the cross-domain registration task, while the traditional SyN~\cite{syn} achieves satisfactory results.}
	\label{fig:intro}
\end{figure}

Medical image registration aligns images acquired at different times, with different modalities, or from different devices to the same spatial domain, aiding doctors in more accurate and effective diagnosis and treatment planning.
While traditional registration methods~\cite{sotiras2013deformable} are applicable in various scenarios, they are often time-consuming for the iterative estimation of the deformation field.
In contrast to traditional methods, deep learning-based registration networks~\cite{vxm,transmorph,transmatch,prpp,pivit,modet,wang2024recursive,rcn,zhu2021joint} greatly accelerate the solving speed of deformation fields.
However, most existing registration networks are constrained to the specific scenarios present in their training sets, leading to a performance degradation when directly applying to new scenarios (see Fig.~\ref{fig:intro}).
Re-training a network for its use in new scenarios entails extra data and time.

Currently, there are limited studies focusing on the domain adaptation of image registration.
Ferrante~\textit{et al}.~\cite{os} investigates the adaptation capability of 2D registration networks in cross-domain scenario, by employing the pre-trained models for one-shot fine-tuning.
In the context of registration, ``one-shot'' refers to fine-tuning a registration network on specific image pairs.
Due to the nature of unsupervised learning, registration networks are allowed to continue unsupervised optimization on new data, aiming to bring registered images closer in terms of similarity.
Subsequently, some studies~\cite{dm1,dm2,wang2024modetv2} follow this concept and directly apply the one-shot learning strategy for fine-tuning registration networks to attain better accuracy.
However, all aforementioned studies only use conventional networks to explore the one-shot learning for single-/cross-domain registration, yet not design network architectures tailored specifically for this strategy.

By rethinking traditional registration methods, it can be summarized the registration algorithms involve three main components: the matching criterion, deformation model, and optimization method.
Among which, the deformation model and optimization method have been extensively studied in recent deep learning-based methods.
However, the matching criterion (i.e., similarity measure), which is closely related to the features extracted by the encoder, has not been thoughtfully studied.
We argue that the powerful and representative encoder could make the registration network more generalized and reusable for cross-domain registration.

In this study, we propose a novel encoder component to explicitly represent matching criteria for the image registration.
Experiments on three public magnetic resonance imaging (MRI) datasets (including brain, abdomen, cardiac images) demonstrate that our method not only has favorable registration accuracy, but also with satisfactory generalization performance, which can be adapted to different domain through simple one-shot learning strategy.
To sum up, our primary contributions are as follows:
\begin{itemize}
	\item[$\bullet$] We design a registration-oriented encoder to explicitly model the matching criteria of image features and structural features, which is beneficial to boost registration accuracy and adaptability.
	\item[$\bullet$] We propose a general feature encoder (Encoder-G), which effectively captures comprehensive medical image features thus enhancing the generalization capability.
	\item[$\bullet$] We propose a structural feature encoder (Encoder-S), which encodes the local self-similarity into the global information, thereby enhancing the structural representation.
	Additionally, by updating Encoder-S using one-shot learning, the model can be effectively adapted to different domains.
\end{itemize}

\section{Method}
\subsection{Network Overview}

\begin{figure}[t]
	\includegraphics[width=\textwidth]{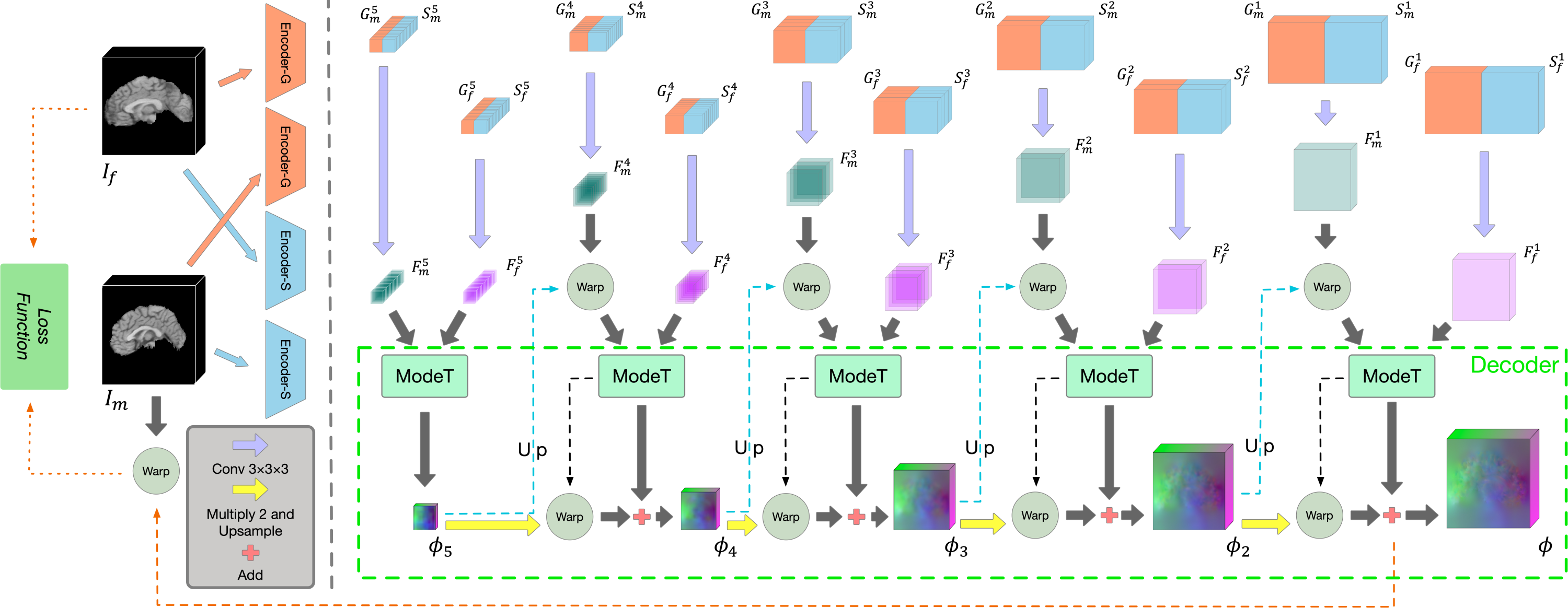}
	\caption{The proposed registration network consists of the registration-oriented Encoder-G and Encoder-S, and the motion decomposition Transformer (ModeT)~\cite{modet} decoder.}
	\label{fig:framework}
\end{figure}

As shown in Fig.~\ref{fig:framework}, the proposed registration network mainly consists of the registration-oriented encoder (Encoder-G and Encoder-S), and the motion decomposition Transformer (ModeT)~\cite{modet} decoder.
Note that the design of decoder is not the focus of this study, thus we directly use latest cutting-edge decoder ModeT.
Encoder-G and Encoder-S are the pyramid structure and each contains two weight-sharing encoders (details are shown in Figs.~\ref{fig:encoderG} and~\ref{fig:encoderS}).
Encoder-G focuses on extracting more generalized image features, while Encoder-S focuses on structural information.
Encoder-G uses a series of convolutional blocks to extract features from both the moving image $I_m\in\mathbb{R}^{H \times W \times L}$ and the fixed image $I_f\in\mathbb{R}^{H \times W \times L}$, resulting in two sets of hierarchical features \{$G^1_m$, $G^2_m$, $G^3_m$, $G^4_m$, $G^5_m$\} and \{$G^1_f$, $G^2_f$, $G^3_f$, $G^4_f$, $G^5_f$\}.
And Encoder-S extracts the features from $I_m$ and $I_f$ defined as \{$S^1_m$, $S^2_m$, $S^3_m$, $S^4_m$, $S^5_m$\} and \{$S^1_f$, $S^2_f$, $S^3_f$, $S^4_f$, $S^5_f$\}.
The $H$, $W$, $L$ denote the image size in height, width and length, respectively.

In the decoder section, we utilize ModeT~\cite{modet} to interpret the feature maps and derive the deformation fields.
Initially, we separately input ($G^5_f$, $S^5_f$) and ($G^5_m$, $S^5_m$) into convolutional layers to perform feature fusion generating $F^5_f$ and $F^5_m$.
The fused features are then used as the input for ModeT to obtain the deformation field $\phi_5$, which is utilized to deform the next-level fused features $F^4_m$ as the input for the next-level ModeT, and so forth to derive the final deformation field $\phi$.
Finally, we employ $\phi$ to deform $I_m$ to obtain the registered image.
The detailed training strategy and training loss are described in Section~\ref{sec:train}.

\subsection{General Feature Encoder}

\begin{figure}[t]
	\centering
	\includegraphics[width=\textwidth]{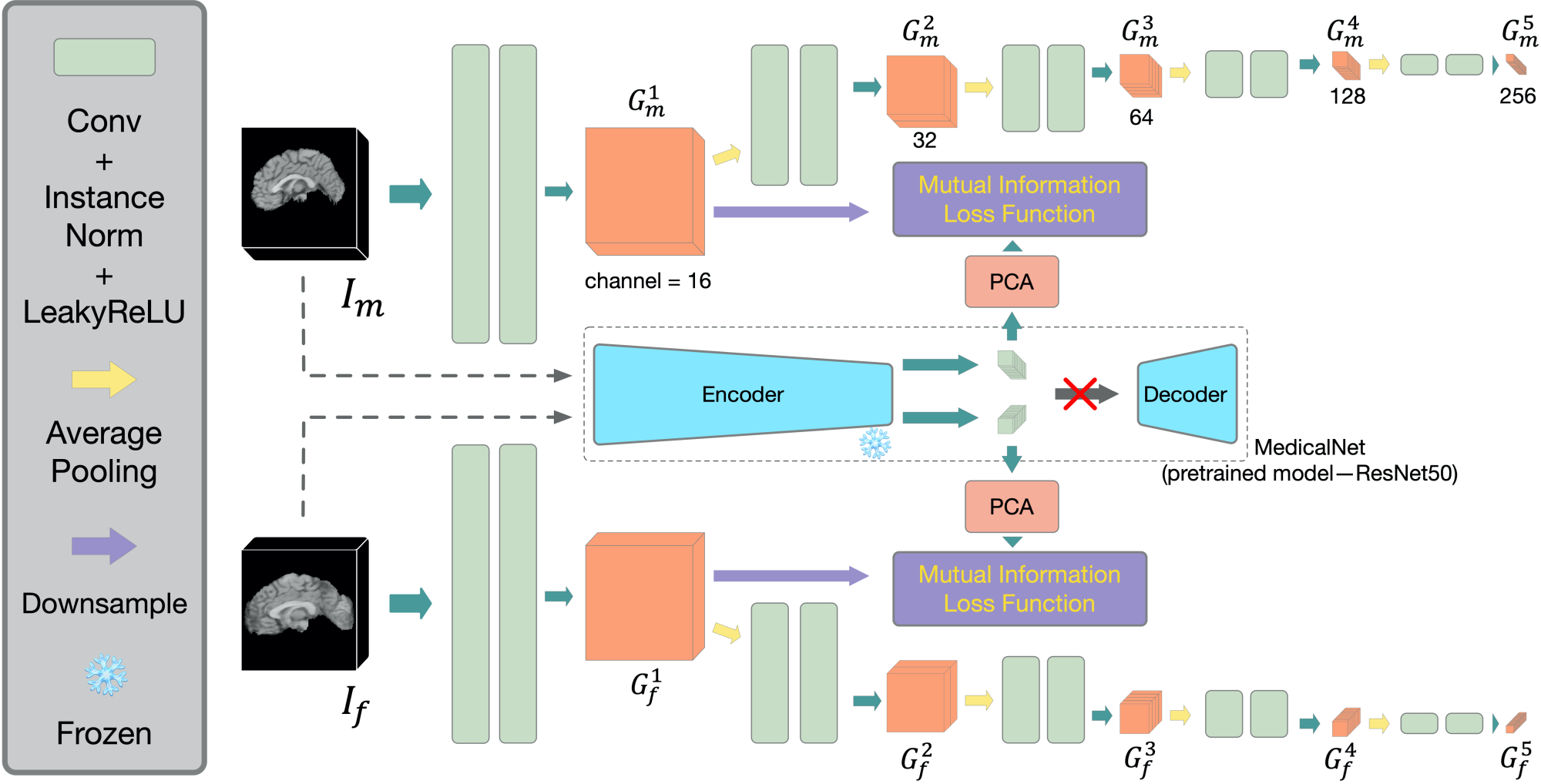}
	\caption{General feature encoder (Encoder-G).}
	\label{fig:encoderG}
\end{figure}

As illustrated in Fig.~\ref{fig:encoderG}, Encoder-G is proposed for extracting general medical image features.
Within each encoder unit, we employ two convolution blocks, each comprising a 3$\times$3$\times$3 convolution layer, an instance normalization layer~\cite{instancenorm}, and a LeakyReLU activation layer.
We utilize the encoder of MedicalNet~\cite{medicalnet} to extract image features and employ principal components analysis (PCA)~\cite{pca} to reduce dimension to the same number of channels as the features extracted by the first layer of the Encoder-G.
We then compute the mutual information~\cite{mutual} with the features generated by the MedicalNet and the features extracted by the first layer of Encoder-G, boosting the capability of Encoder-G to learn more comprehensive medical image information in line with the MedicalNet.

\subsection{Structural Feature Encoder}

\begin{figure}[t]
	\includegraphics[width=\textwidth]{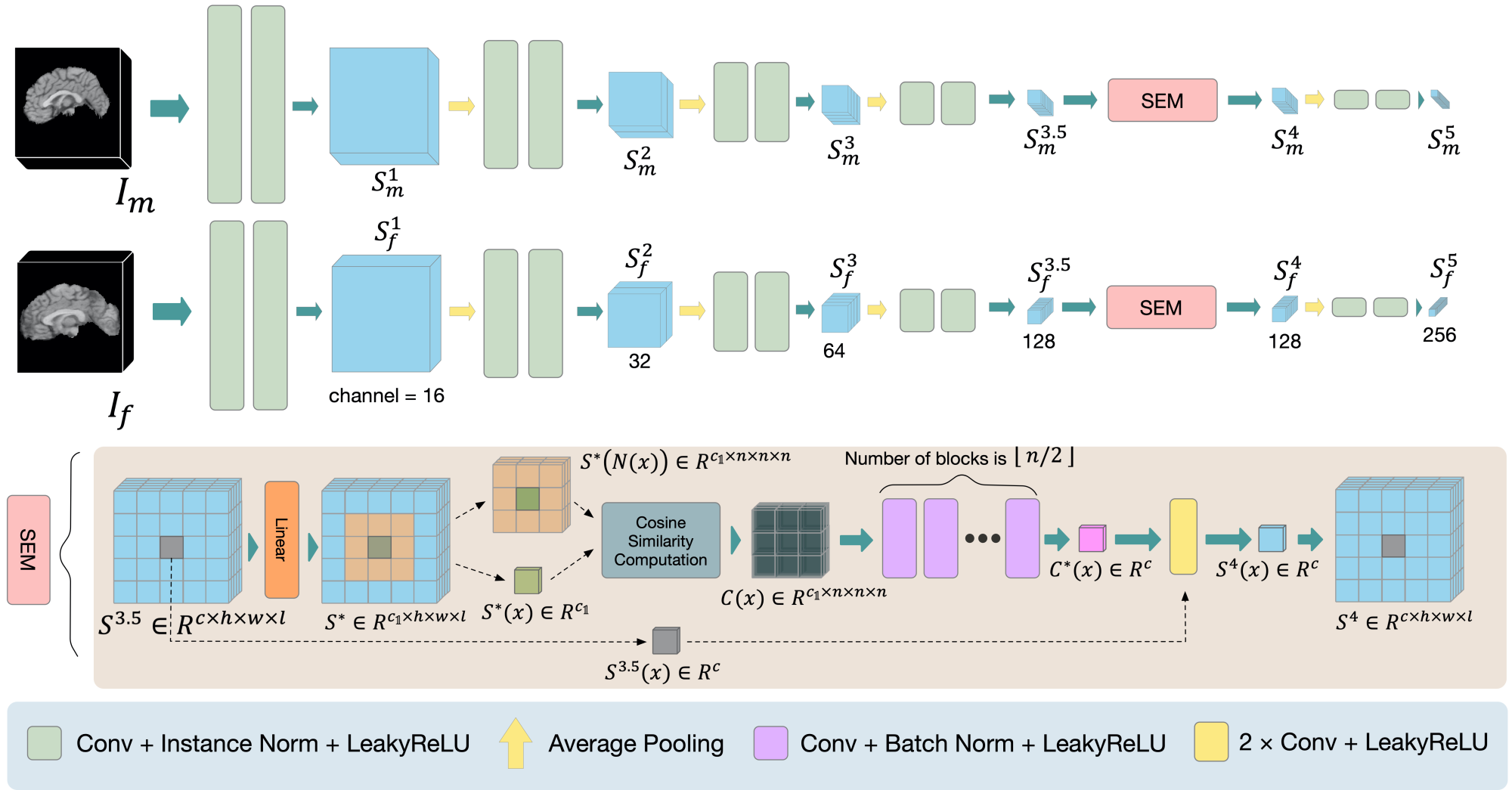}
	\caption{Structural feature encoder (Encoder-S) and the structural embedding module (SEM).
	$h$, $w$, $l$ and $c$ represent the height, width, length and channel number of the features, respectively.
	$N(x)$ denotes the $n \times n \times n$ neighborhood for voxel $x$.}
	\label{fig:encoderS}
\end{figure}

As shown in Fig.~\ref{fig:encoderS}, we devise the Encoder-S with the embedded structural embedding module (SEM) to encode the structural self-similarity into the global representation.
The SEM initially reduces the feature dimensionality through the linear layer, thereby reducing the computation burden.
It then calculates the self-similarity between each voxel $x$ and its local neighboring regions with size of $n \times n \times n$ using cosine similarity：
\begin{equation}
C(c',x,d) = max(0, \frac{S^*(c',x) \cdot S^*(c',x+d)}{||S^*(c',x)|| \cdot ||S^*(c',x + d)||}),
\label{equcosine} 
\end{equation}
where $c' \in [1, c_\mathds{1}]$ is index of channel dimension and $d \in [-D,D] \times [-D,D] \times [-D,D]$ represents the relative position of voxel x in the surrounding region with size of $n$. $D=(n-1)/2$.
The resulting self-similarity matrix is encoded into a structural feature vector of the same dimension as the input of self-similarity computation through multiple convolution layers.
The structural feature vectors are then sequentially summed and convolved with the input to produce the final enhanced structural feature maps.

\subsection{Training Strategy}
\label{sec:train}
We adopt a three-stage training strategy.
\textbf{Stage 1}: optimize Encoder-G and decoder.
The feature convolution fusion operations (purple arrows) in Fig.~\ref{fig:framework} are disabled, which means we have $F_m$ equivalent to $G_m$, and $F_f$ equivalent to $G_f$.
The training loss is as follows:
\begin{equation}
\mathcal{L}_{\operatorname{stage1}} = \mathcal{L}_{\operatorname{ncc}}(I_f, I_m \circ \phi) + \lambda\mathcal{L}_{\operatorname{reg}}(\phi) + \mathcal{L}_{\operatorname{mi}},
\label{equloss} 
\end{equation}
where $\mathcal{L}_{\operatorname{ncc}}$, $\mathcal{L}_{\operatorname{reg}}$, and $\mathcal{L}_{\operatorname{mi}}$ represent the conventional normalized cross correlation~\cite{rao2014application}, deformation regularization~\cite{modet} and mutual information~\cite{mutual}, respectively.
The operation $\circ$ denotes warping and $\lambda$ is a weighting factor.
Note that the $\mathcal{L}_{\operatorname{mi}}$ is only computed in the last twenty percent of the training epochs.
\textbf{Stage 2}: optimize Encoder-S and decoder, freeze Encoder-G.
This means we have $F_m$ equivalent to $S_m$, and $F_f$ equivalent to $S_f$.
The training loss is as follows:
\begin{equation}
\mathcal{L}_{\operatorname{stage2}} = \mathcal{L}_{\operatorname{ssim}}(I_f, I_m \circ \phi) + \lambda\mathcal{L}_{\operatorname{reg}}(\phi),
\label{equloss2} 
\end{equation}
where $\mathcal{L}_{\operatorname{ssim}}$ calculates the structural similarity index measure~\cite{mahmood2020exploring}.
\textbf{Stage 3}: freeze Encoder-G and Encoder-S, optimize the convolutional fusion layer and the decoder.
%In order to enhance the collaboration between Encoder-G and Encoder-S, Encoder-G and Encoder-S are frozen, and features of the same size and from the same image extracted by these two parts will be synthesized into a new feature through a $3\times3\times3$ convolution module. The new feature will be deformed by the deformation field generated by the previous decoder and input to the next decoder.
The training loss is as follows:
\begin{equation}
\mathcal{L}_{\operatorname{stage3}} = \mathcal{L}_{\operatorname{ncc}}(I_f, I_m \circ \phi) + \lambda\mathcal{L}_{\operatorname{reg}}(\phi).
\label{equloss3} 
\end{equation}

\subsection{One-shot Learning for Inference Stage}
The one-shot learning is employed for the adaptation of cross-domain registration.
In the inference stage, we freeze Encoder-G and activate the fusion module, decoder, and Encoder-S.
For each image pair to be inferred, we retrain the model for 5, 10, and 20 iterations using $\mathcal{L}_{\operatorname{stage2}}$, then perform inference immediately after training.
Note that for the next image pair, we reload the initial model then repeat the one-shot learning.

\section{Experiments}
\subsubsection{Datasets.}
Experiments were conducted on three public magnetic resonance imaging (MRI) datasets,
including brain dataset LPBA40~\cite{dataset_lpba},
abdomen dataset AbdMR~\cite{dataset_abdo},
and cardiac dataset~\cite{dataset_cardiac}.
\textbf{The LPBA40 dataset} contains 40 brain MRI scans, each with 54 labeled region-of-interests (ROIs).
All MRI scans were resampled to the size of $160\times192\times160$.
30 scans (30$\times$29 pairs) were employed for training and 10 scans (10$\times$9 pairs) were used for testing.
\textbf{The AbdMR dataset} contains abdominal MRI scans with 4 annotated organs.
Each MRI scan has the size of $192\times160\times192$.
38 scans (38$\times$37 pairs) were used for training and 10 scans (10$\times$9 pairs) were used for testing. 
\textbf{The Cardiac dataset} contains cardiac MRI scans with 3 annotated structures.
Each MRI scan has the size of $128\times96\times128$.
255 scans (255$\times$254 pairs) were used for training and 30 scans (30$\times$29 pairs) were used for testing.

\subsubsection{Implementation Details.}
%We utilize Motion Decomposition Transformer(ModeT) as the central element of our Decoder for predicting feature motion patterns. Recognizing that the Competitive Weighting Module, employed in the original paper to fuse motion patterns, was overly memory-intensive, we opted for a more efficient solution by replacing it with a simple convolutional layer.
We deployed our registration network using PyTorch on a NVIDIA Tesla V100 GPU equipped with 32GB memory.
We used the default setting of ModeT~\cite{modet}.
%The number of attention heads $\epsilon$ in ModeT were set to 8, 4, 2, 1, 1, from deep to shallow, and each head was equipped with 6 channels.
The parameter $n$ in SEM was set to 7.
The weighting factor $\lambda$ was set to 1 for brain images, and 0.5 for abdomen and cardiac images.
The training was conducted using Adam optimizer with a learning rate of 0.0001.
The batch size was 1.

\subsubsection{Comparison Methods and Evaluation Metrics.}
To demonstrate the efficacy of the proposed method, we compared several cutting-edge registration methods:
(1) SyN~\cite{syn}: a classical iterative registration method.
(2) Im2Grid (I2G)~\cite{im2grid}: a coordinate translator CNN-based registration model.
(3) VoxelMorph (VM)~\cite{vxm}: a popular CNN-based single-stage registration model.
(4) Dual-PRNet++ (DPR++)~\cite{prpp}: a CNN-based pyramid registration model.
(5) TransMorph (TMP)~\cite{transmorph}: a Transformer-based registration model.
(6) TransMatch (TMT)~\cite{transmatch}: a dual-stream Transformer-based registration model.

We employed Dice similarity coefficient (DSC) and average symmetric surface distance (ASSD) to evaluate the registration accuracy.
The DSC quantifies the region-based similarity between the fixed and registered images, while the ASSD measures the surface-based similarity.
Higher DSC and lower ASSD values indicate better registration.
% The Jacobian determinant of the deformation field ($|J(\phi)|<0$)~\cite{modet} assesses the smoothness of the deformation. 
%For visualization, we only computed DSC for cross-domain tasks and represented the results graphically.

\begin{table}[t]
	\centering
	\caption{Quantitative results of different methods on three public datasets}
	\begin{tabular}{l|c|c|c|c|c|c}
		\toprule  
		%& AbdMR &&  Cardiac&& LPBA40& \\
		& \multicolumn{2}{c|}{AbdMR (4 ROIs)} & \multicolumn{2}{c|}{Cardiac (3 ROIs)} & \multicolumn{2}{c}{LPBA40 (54 ROIs)} \\
		\cmidrule{2-7}
		%\midrule
		&  DSC (\%)&  ASSD&   DSC (\%)&  ASSD&  DSC (\%)&  ASSD
		\\
		\midrule
		Initial&  48.5$\pm$10.2&  14.13$\pm$4.30&   37.4$\pm$13.4&  8.23$\pm$2.93&  53.7$\pm$4.8&  2.49$\pm$0.33
		\\
		\midrule
		SyN~\cite{syn}&  63.2$\pm$15.7$^*$&  12.06$\pm$7.80$^*$ &  53.7$\pm$14.8$^*$&  5.67$\pm$2.89$^*$&  70.4$\pm$1.8$^*$&  1.46$\pm$0.13$^*$
		\\
		I2G~\cite{im2grid}&  49.6$\pm$10.0$^*$&  14.27$\pm$4.29$^*$&  36.9$\pm$11.2$^*$&  8.87$\pm$2.79$^*$&  69.4$\pm$1.8$^*$&  1.49$\pm$0.14$^*$
		\\
		VM~\cite{vxm}&  61.7$\pm$10.5$^*$&  11.51$\pm$4.53$^*$&  48.5$\pm$13.9$^*$&  6.08$\pm$2.56$^*$&  62.7$\pm$3.6$^*$&  1.86$\pm$0.24$^*$
		\\
		DPR++~\cite{prpp}&  65.5$\pm$10.8$^*$&  10.29$\pm$4.50$^*$&  47.7$\pm$15.0$^*$&  6.41$\pm$2.85$^*$&  68.1$\pm$2.5$^*$&  1.58$\pm$0.18$^*$
		\\
		TMT~\cite{transmatch}&  64.7$\pm$9.5$^*$&  10.65$\pm$4.33$^*$&  47.9$\pm$15.0$^*$&  6.38$\pm$2.83$^*$&  65.6$\pm$3.2$^*$&  1.71$\pm$0.22$^*$
		\\
		TMP~\cite{transmorph}&  64.8$\pm$9.4$^*$&  10.66$\pm$4.28$^*$&  48.0$\pm$15.0$^*$&  6.34$\pm$2.82$^*$&  65.8$\pm$3.3$^*$&  1.71$\pm$0.22$^*$
		\\
		\midrule
		Ours&  \textbf{68.9$\pm$11.8}&  \textbf{9.75$\pm$5.16}&  \textbf{65.2$\pm$12.4}&  \textbf{3.82$\pm$2.20}&  \textbf{72.8$\pm$1.3}&  \textbf{1.30$\pm$0.09}
		\\
		\bottomrule
	\end{tabular}
	\label{tab:my_label}
\end{table}

\begin{figure}[t]
	\includegraphics[width=\textwidth]{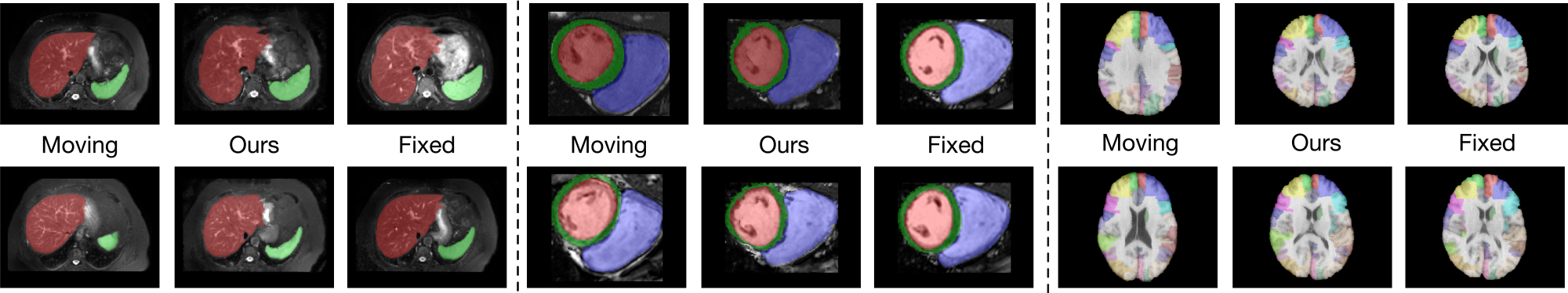}
	\caption{The registration results on different datasets.  The colored regions indicate the masks of different anatomical structures.}
	\label{fig:results}
\end{figure}

\subsubsection{Single-Domain Results.}
The quantitative results of different methods on three datasets are listed in Table~\ref{tab:my_label}.
In this Table, the symbol $^*$ indicates that the DSC/ASSD results are statistically different from ours (Wilcoxon tests, $p < 0.05$).
It is suggested that our method significantly outperformed the other compared methods for the single-domain task on these three datasets.
Notably, regarding the DSC results, our method surpassed the second-best methods by 3.4\%, 11.5\% and 2.4\% on AbdMR, Cardiac and LPBA datasets, respectively.
Additionally, our method consistently achieved low percentages of non-positive Jacobian determinant voxels ($\%|J(\phi)|\leq 0$) on the three datasets, with values of $<$2\%, $<$1\%, and $<$0.002\%, respectively.
This indicates that our method generated smooth deformation fields.
%From the table, we observe that registering 3D Cardiac data is particularly challenging. However, our network performs exceptionally well in completing this task.
Fig.~\ref{fig:results} further visualizes our registration results on different datasets.
Our method accurately registered corresponding anatomical structures.

\begin{figure}[t]
	\includegraphics[width=\textwidth]{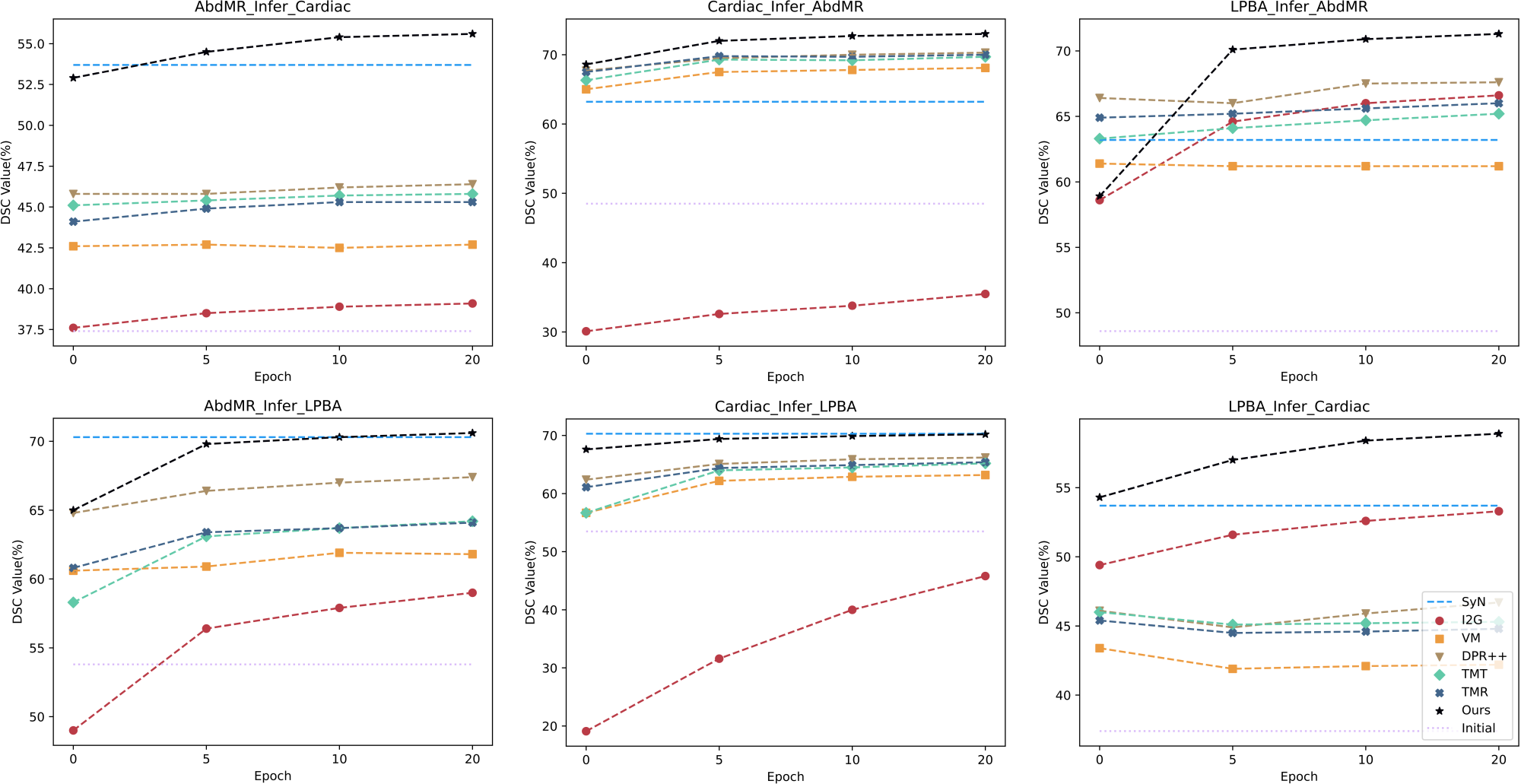}
	\caption{DSC results of different methods using one-shot learning for cross-domain registration.
	``A\_Infer\_B'' refers to the model trained on dataset A infers dataset B.}
	\label{fig:oneshot}
\end{figure}

\subsubsection{Cross-Domain Results.}
Regarding the cross-domain registration, Fig.~\ref{fig:oneshot} illustrates the improvement achieved by the one-shot learning strategy across different networks.
By observing the results in Fig.~\ref{fig:oneshot}, we can have the following findings:
(1) Our method had overall the best generalization capability among all compared deep learning-based methods.
Without one-shot optimization, our method outperformed other deep networks in five of six cross-domain scenarios, and even beat SyN in two scenarios.
(2) By conducting 20 iterations of one-shot optimization, our method successfully surpassed SyN in all six cross-domain scenarios.
(3) Actually by using only 5 iterations of one-shot optimization, our method already had better or very close performance compared to SyN did.
This indicates our method could be effectively adapted to different domains.
%We found that conducting more than 5 iterations of optimization is less cost-effective, leading us to conclude that 5 iterations of one-shot optimization represent the most effective strategy.

On average, our method required 17.3 seconds to register a pair of volumes in 5-iterations one-shot optimization.
The time spent included model loading, loss function calculation, model parameters updating and inference.
In contrast, SyN took 114.7 seconds for registering the same pair of volumes.

\section{Conclusion}
Our main motivation is to leverage a registration-oriented encoder to effectively model the matching criteria of image features and structural features and therefore boosting the registration performance, especially for the cross-domain task.
To this end, we devise the general feature encoder (Encoder-G) to extract generic medical image information, as well as the structural feature encoder (Encoder-S) to optimize the representation of the structural information.
The Encoder-S can be optimized using one-shot learning during inference stage, for the purpose of domain adaptation.
The proposed method has been tested using three different MRI datasets.
Our method yields favorable results in the single-domain task while ensuring improved generalization and adaptation performance in the cross-domain task.
Nonetheless, our method is still undergoing refinement, and we aim to further enhance its speed for broader clinical applications in the future.

\section*{Acknowledgements}
This work was supported in part by the National Natural Science Foundation of China under Grant 62071305,
in part by the Guangdong-Hong Kong Joint Funding for Technology and Innovation under Grant 2023A0505010021,
and in part by the Guangdong Basic and Applied Basic Research Foundation under Grant 2022A1515011241.

\bibliographystyle{splncs04}
\bibliography{bib2}

\end{CJK}   
\end{document}